# Leveraging Industry 4.0 - Deep Learning, Surrogate Model and Transfer Learning with Uncertainty Quantification Incorporated into Digital Twin for Nuclear System

M. Rahman[1], Abid Khan[2], Sayeed Anowar[3], Md Al-Imran[4], Richa Verma[5], Dinesh Kumar[6], Kazuma Kobayashi[7], Syed Alam[7]

[1]Department of Mechanical Engineering, Bangladesh University of Engineering and Technology, Dhaka 1000, Bangladesh

[2]Institute of Nuclear Power Engineering, Bangladesh University of Engineering and Technology, Dhaka-1000, Bangladesh

[3]Department of Industrial and Production Engineering, Jashore University of Science and Technology, Jashore 7408, Bangladesh

[4]Department of Electrical and Electronic Engineering, Bangladesh University of Engineering and Technology, Dhaka 1000, Bangladesh

[5]Department of Electrical Engineering, Indian Institute of Technology Delhi, Delhi 110016, India

[6]Department of Mechanical Engineering, University of Bristol, Bristol BS8 1TR, UK

[7]Department of Nuclear Engineering and Radiation Science, Missouri University of Science and Technology, Rolla, MO 65409, USA

*Corresponding author: Syed Alam (alams@mst.edu)

## Abstract

Industry 4.0 targets the conversion of the traditional industries into intelligent ones through technological revolution. This revolution is only possible through innovation, optimization, interconnection, and rapid decision-making capability. Numerical models are believed to be the key components of Industry 4.0, facilitating quick decision-making through simulations instead of costly experiments. However, numerical investigation of precise, high-fidelity models for optimization or decision-making is usually time-consuming and computationally expensive. In such instances, data-driven surrogate models are excellent substitutes for fast computational analysis and the probabilistic prediction of the output parameter for new input parameters. The emergence of Internet of Things (IoT) and Machine Learning (ML) has made the concept of surrogate modeling even more viable. However, these surrogate models contain intrinsic uncertainties, originate from modeling defects, or both. These uncertainties, if not quantified and minimized, can produce a skewed result. Therefore, proper implementation of uncertainty quantification techniques is crucial during optimization, cost reduction, or safety enhancement processes analysis. This chapter begins with a brief overview of the concept of *surrogate modeling, transfer learning, IoT and digital twins*. After that, a detailed overview of uncertainties, uncertainty quantification frameworks, and specifics of uncertainty quantification methodologies for a surrogate model linked to a digital twin is presented. Finally, the use of uncertainty quantification approaches in the nuclear industry has been addressed.

## 1. Introduction

The concept of Industry 4.0 has attracted an enormous amount of attention of scientific community in the last few years. This concept is emerged from the rapid changes in global industrial landscape due to technological innovations and advancements in the past decade. Just as First Industrial Revolution can be marked by the invention of steam engine, Second Industrial Revolution by the replacement of steam by chemical and electrical energy, and Third Industrial Revolution by the invention of microchips, the Fourth Industrial Revolution is expected to be introduce the widespread interaction of digital and physical systems through Internet of Things (IoT) and Augmented reality for increased productivity (Pereira and Romero, 2017). Because of the immense potential of socio-economic development, this impact of Industry 4.0 is rigorously being researched by the scientists and academicians alike.

While there are many concepts and technologies incorporated within the domain of Industry 4.0, a lot of them are directly linked to computer simulation. The first one that comes in mind is cyber-physical system (CPS). CPS includes machines, systems and processes that can exchange information autonomously and take decisions accordingly. CPS is believed to be the core of smart factory (Gunal, 2019). In a CPS, the physical system can be regulated by a digital replica of the system known as "Digital Twin" (DT). In reality, it is nothing but a simulation of the actual system that it replicates. The main role of DT is to simulate the outcome of a decision in the virtual world and help making decisions in the real world. Since there is data exchange between DT and the real world through sensors, the simulation performed are "real-world aware" (Gunal, 2019). Agmented Reality (AR) and Virtual Reality (VR) are two other technologies that are reliant on computer simulations. Although they are optional in some industries, they will be compulsory in future industries (Gunal, 2019). There are numerous other parts of Industry 4.0 such as vertical and horizontal system integration, hybrid modeling, cloud computing, additive manufacturing and product design, etc. that will make extensive use of computer simulation. Thus, simulation is considered as an essential tool for making the Fourth Industrial Revolution happen.

Numerical methods are a long-standing problem-solving route, as old as the Egyptian Rhind papyrus (c. 1650 BC), describing a root-finding method of a simple equation. However, the physics-based modern numerical analysis or computer simulations that simulate physical systems using mathematical concepts and language essentially began with studying sources and rounding the errors by Von Neumann and Goldstine (1947). The study of numerical modeling has significantly progressed since then and opened new ways to investigate various problems that would be extremely difficult, if not impossible, to investigate experimentally (Avvakumov *et al.*, 2018; Galati and Iuliano, 2018; Abedini and Zhang, 2021). However, most of these models require heavy computation to produce results with acceptable accuracy, for they simulate the physics behind the problems. In many cases, the simulations are so computationally expensive that it becomes rather impractical to utilize them for decision-making and robust optimization problems. A new inclusion to the fleet of computational modeling, the data-driven surrogate modeling, has the solution to this problem. Although initial training of a surrogate model can be computationally expensive and requires a significant amount of precise data, it can produce high fidelity in significantly less time once it is properly trained (Jiang, Zhou and Shao, 2020).

However, despite having a vast library of algorithms, substantial computational power, and enough training data, both surrogate models and numerical models cannot accurately predict a real-world problem in many cases. Thus, uncertainty quantification, which measures uncertainty refers to the deviations of model output from actual data, is equally important as the model.

In this chapter, the potential role of Industry 4.0 in the flourish of the futuristic nuclear industry is described in brief. The, the concept of surrogate modeling is introduced. After that, the general idea of uncertainty, its classifications, and methods for quantifying uncertainty have been explained. Later, different uncertainty quantification methods applied to the surrogate models have been further discussed. Finally, a literature review was undertaken to see how uncertainty quantification in surrogate models is employed in nuclear applications.

## 2. Industry 4.0 in the Context of Nuclear System

Although the Third Industrial Revolution introduced the concept of industrial automation, Industry 4.0 will be the one to take it to a whole new level. Several advanced technologies such as Cloud computation, artificial intelligence (AI), IoT, etc. is expected to take over to achieve a smart industrial network. Since the complex interconnected systems realized in the vision of Industry 4.0 consists of smart sensors, devices, robots, etc., the electricity consumption is expected to grow exponentially. One of the key challenges for Industry 4.0 is to find green energy sources that can meet the energy demand. This is where nuclear energy comes in. It is a proven and reliable source of energy that can propel the concept of Fourth Industrial Revolution towards real-life implementation (Lu *et al.*, 2020).

While supplying the low-carbon energy for Industry 4.0, Nuclear Power Plants (NPPs) will actually do the favor to themselves. It is the concept of Industry 4.0 that can help this power industry regain the acceptance it lost among the mass population due to some serious accidents like Chernobyl disaster, Fukushima disaster, etc (Lu *et al.*, 2020). This is because there exists multiple barriers and risks associated with concurrent NPP designs. Due to the complex human-machine-network integration in a nuclear system, faults and failures may occur from multiple directions, making it extremely challenging for an operator to control the NPP on his own. This situation can be overcome with the help of automation technologies assisted by AI (Lu *et al.*, 2020).

With the growing popularity of AI, researchers have proposed utilizing it in a NPP in various manners including cyber-physical system-based security verification (Gawand, Bhattacharjee and Roy, 2017; Tripathi *et al.*, 2021), in-core fuel management (Jiang *et al.*, 2006), real-time core monitoring (Xia, Li and Liu, 2014), core parameter estimation (Hedayat *et al.*, 2009), fault detection (Yao *et al.*, 2022; Yao, Wang and Xie, 2022), etc. A concept that is in harmony with the vision of Industry 4.0 is that of Nuclear Energy 5.0, which blends industrial blockchain, automation and AI together (Wang *et al.*, 2018). A more recent concept is Nuclear Power Plant Human-Cyber-Physical System (NPPHCPS) proposed by Lu *et al.* (2020). In this conceptual framework shown in Fig.1, the physical system layer at the bottom deals transfers data to the cyber system layer for AI-based processing. This layer comes in aid of the human operator layer lying on the top through autonomous control, intelligent monitoring and error minimization (Lu *et al.*, 2020).

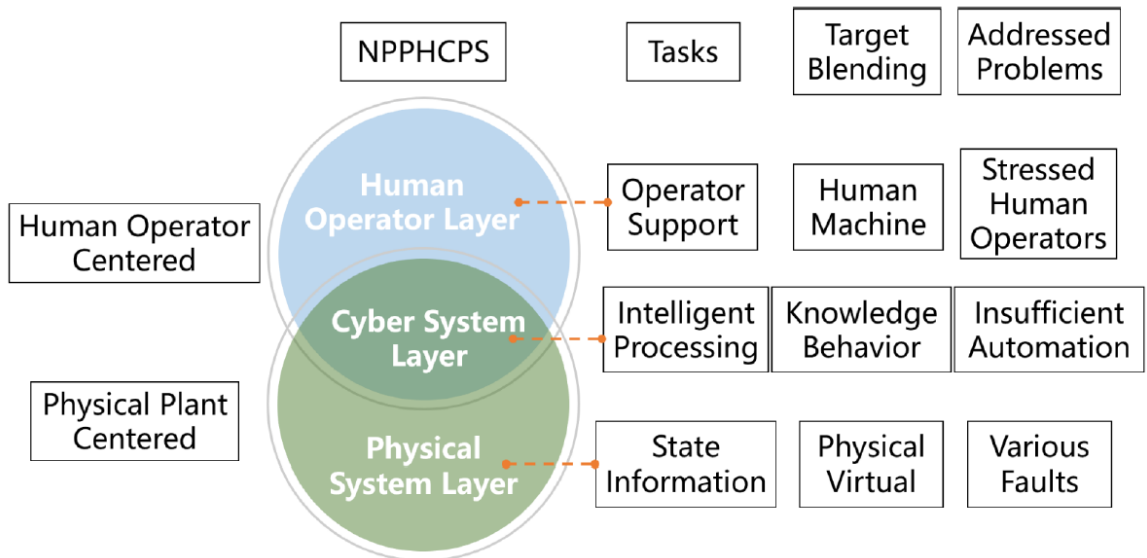

**Fig.1.** Conceptual framework of NPPHCPS (Lu *et al.*, 2020)

The NPPHCPS proposed by Lu *et al.* (2020) is essentially a deep learning framework. For practical implementation of this framework, generation and processing of a large volume of data will be required. As a result, High Performance Computing (HPC) will be essential in most cases. In order to reduce computational expenses, concepts like surrogate modeling and transfer learning may play a crucial role.

## 3. Surrogate Modeling, Deep Learning & Transfer Learning

The surrogate model is a data-driven model developed to predict the behaviour of a physical system without the need of computationally expensive simulations (Sobester, Forrester and Keane, 2008; Jiang, Zhou and Shao, 2020). These statistical models utilize machine learning (ML) and AI to derive approximation functions that can represent the actual physical system with good accuracy. The main advantage of using a ML-trained surrogate model over direct simulation is that they require less time to generate a large volume of data for performing further research and development (R&D) purpose (Davis, Cremaschi and Eden, 2018). The concept of surrogate modeling is quite straightforward; data-driven fitting functions, known as approximation functions, are derived through ML-based training process. An active learning process is employed to reduce the number of data points necessary to train the model properly (Sobester, Forrester and Keane, 2008). In this process, special learning functions are utilized to identify new sample points for better training of the derived approximation function. After that, simulations are performed to generate data for re-training the model. The process is repeated until the approximation function is sufficiently accurate (Sobester, Forrester and Keane, 2008).

Since their introduction, the surrogate models have been used in the field of water resource (Razavi, Tolson and Burn, 2012), aerodynamic design (Sun and Wang, 2019; Yondo *et al.*, 2019), chemical process engineering (McBride and Sundmacher, 2019), sustainable building design (Evins, 2013), design and operation of nuclear reactor (Figueroa and Göttsche, 2021; Sobes *et al.*, 2021; Tallman *et al.*, 2021), nuclear waste management (Clement *et al.*, 2021), and so on. ML and AI-based algorithms like least square (Hastie, Tibshirani and Friedman, 2001), artificial neural network (ANN) (Eason and Cremaschi, 2014), support vector regression (Brereton and Lloyd, 2010), etc., are being used to train a surrogate model.

With the ongoing flourish of the Internet of Things (IoT) and big data, deep learning-based methods are gradually coming into limelight. Methods like automatic deep encoder–decoder, deep belief networks (DBNs) and deep convolutional neural networks (CNNs) are becoming popular among researchers for developing surrogate models (Mandal *et al.*, 2017; Peng *et al.*, 2018; Saeed *et al.*, 2020). Special interest is observed for developing intelligent fault detection and diagnosis (FDD) methods for accident tolerant fuels (Yao *et al.*, 2022; Yao, Wang and Xie, 2022). Nowadays, researchers are inclining towards CNN rather than using feed-forward neural networks (FFNN) because of its substantial reduction in total number of parameters of the network (Emmert-Streib *et al.*, 2020). In contrast with traditional neural networks, the neurons in the CNN are arranged in three dimensions (Yao *et al.*, 2022), as is shown in Fig.2.

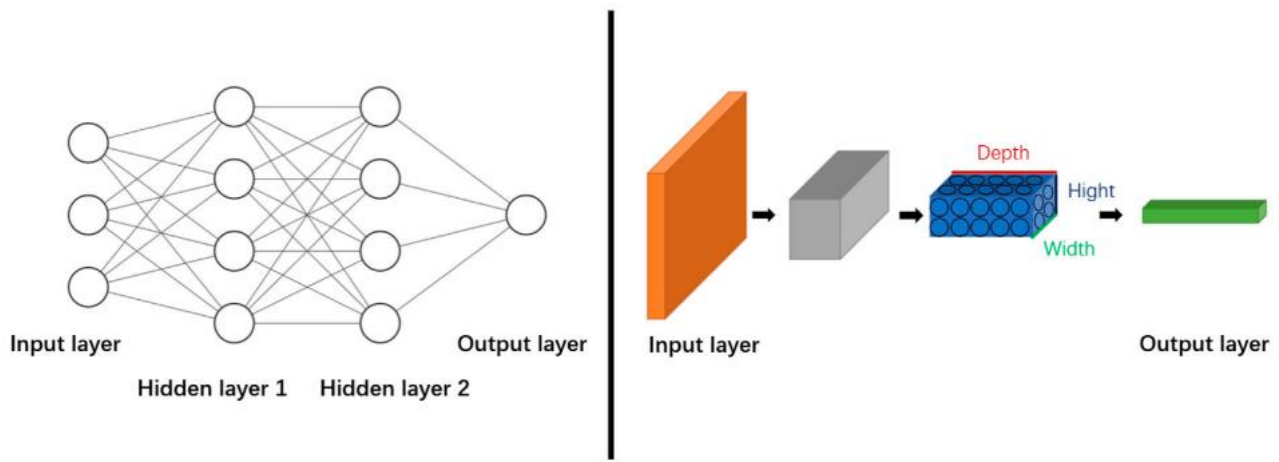

**Fig.2.** Traditional FFNN vs. CNN (Yao *et al.*, 2022)

The neurons of a CNN are not fully connected like traditional NNs; they are only connected to a small portion of the preceding layer. There are three essential parts of a CNN (Emmert-Streib *et al.*, 2020):

- convolutional layer,
- pooling layer, and
- fully connected layer

The convolution layer is the specific part of CNN. It uses locally connected neurons and shared weights (called kernels) to perform the calculations. Fig.3 presents the structure and weighing method of a typical CNN. From the figure, the local connectivity between the neurons and the weight sharing is easily realized.

The main challenge of developing surrogate models with deep-learning methods for nuclear systems is the lack of adequate training samples for fitting the non-linear relationship (Yao *et al.*, 2022). A nuclear system is, for most of the time, operational under stable condition and it is very difficult to get fault data. As a result, the sample size is too small to properly train the deep learning-based surrogate model (Yao *et al.*, 2022; Yao, Wang and Xie, 2022). To tackle this situation, transfer learning (TL) can be adopted. It is the process of transferring information from a one domain to a different but related domain for improved learning (Weiss, Khoshgoftaar and Wang, 2016).

In the TL approach, a previously trained model is used to train a similar model. In this approach, a portion of the pre-trained network of a model is shared by another model that will be partially trained with new dataset. The pre-trained portion will remain "freeze" while the remaining portion will be updated and finetuned. As a result, the computational efficiency is increased significantly (Weiss, Khoshgoftaar and Wang, 2016). Fig.4 presents a schematic representation of the TL approach. With the expansion of the big data repositories, the TL approach is becoming more and more attractive. It has been successfully implemented in text recognition (Wang and Mahadevan, 2011), image processing (Kulis, Saenko and Darrell, 2011; Li *et al.*, 2013), software defect identification (Nam *et al.*, 2017), and multi-language translation (Zhou *et al.*, 2014), etc.

While TL can be accommodated with any surrogate model utilizing deep learning, TL is especially useful for variable fidelity surrogate models (VFSMs). Generally, a VFSM is developed through information fusion between high-fidelity model (HFM) and low-fidelity model (LFM). To do so, an appropriate bridge function is constructed (Tian *et al.*, 2020, 2021). However, it becomes very difficult to construct the bridge function in most cases due to the non-linear relationship between HFM and LFM, requiring rigorous trial and error (Tian *et al.*, 2020). Instead, a low-fidelity surrogate model (LFSM) is constructed first. Then, and with the help of the HFM information, the actual VFSM is constructed using TL (Tian *et al.*, 2021). TL is also very well-suited for multi-fidelity physics informed neural networks (PINNs), where the low data availability makes most of the conventional ML techniques ineffective (Goswami *et al.*, 2020; Chakraborty, 2021). Using TL, it is possible to train a multi-fidelity PINN more efficiently while preventing overfitting of high-fidelity data (Chakraborty, 2021).

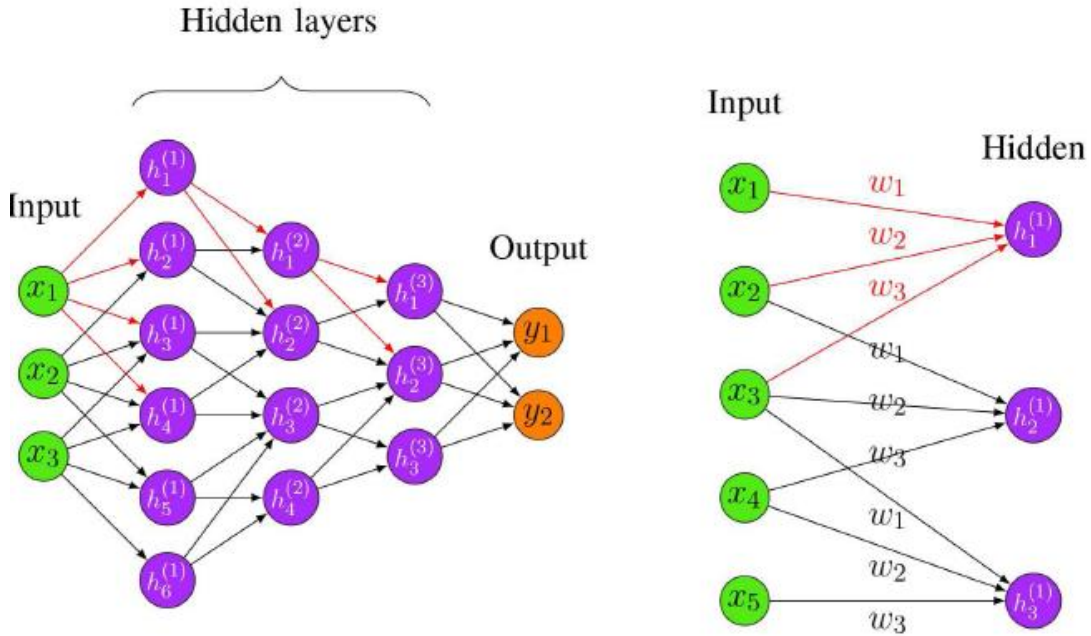

**Fig.3.** Local connectivity and shared weights of a typical CNN (Emmert-Streib *et al.*, 2020)

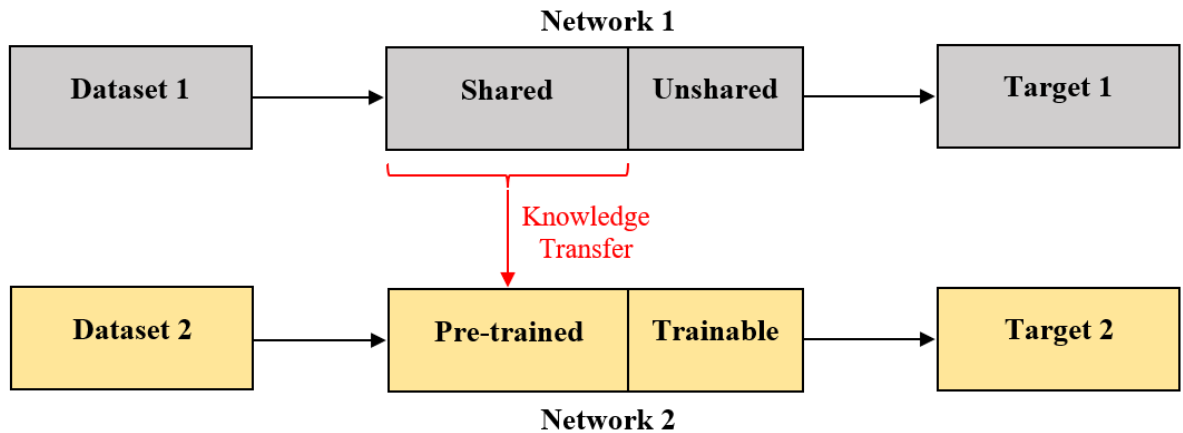

**Fig.4.** A typical transfer learning scheme

Yao *et al.* (2022) proposed utilizing the TL method for fault detection and diagnosis in NPPs. Fig.5 shows the proposed a TL-based fault diagnosis model. There are mainly four steps in the model:

- Construction of the research dataset through existing experimental platforms,
- Pre-setting i.e., training and optimization of the CNN-based model,
- Construction of transferable model through freezing and tuning of some layer parameters, and
- Testing and optimization of proposed TL-based CNN model

According to the study of Yao *et al.* (2022), The proposed TL-based model successfully saved 53.21% of the training time compared to the benchmark CNN model.

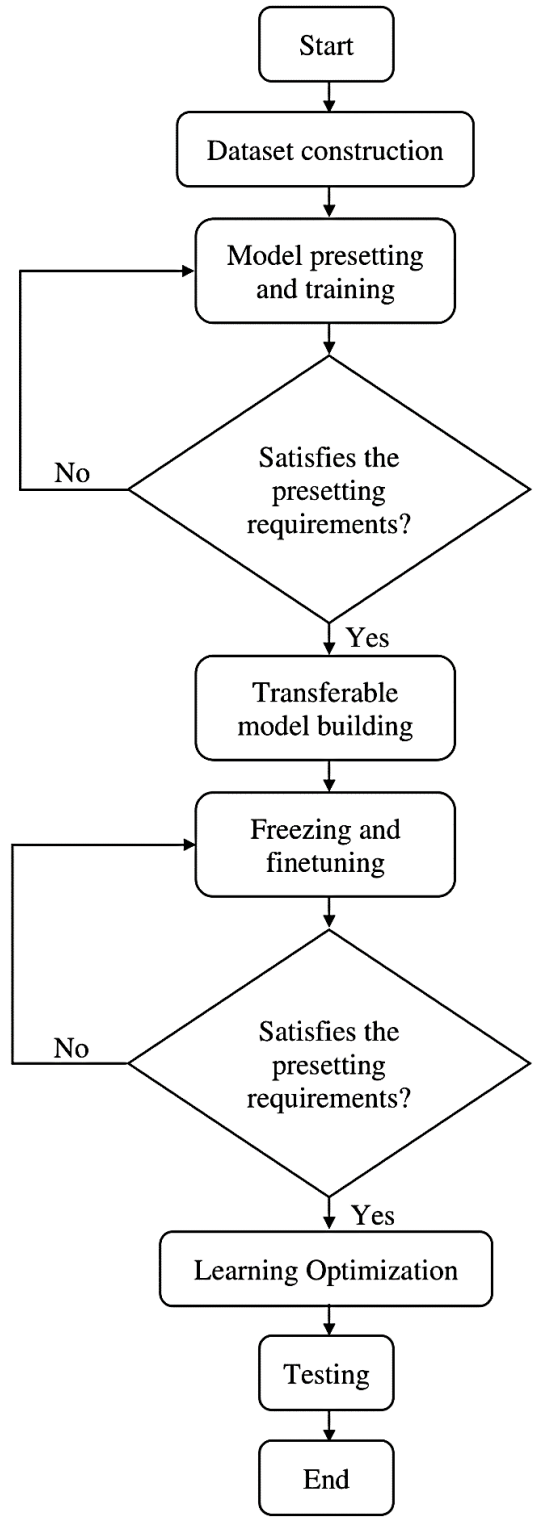

**Fig.5** A deep TL-based fault detection model (Yao *et al.*, 2022).

## 4. Digital Twins

Digital twins (DTs) are virtual mimics of a physical system that utilize real-time data along with simulation results, ML and user input, and help make good decisions or predictions. This technology is considered a crucial and innovative method in various engineering and industrial sectors such as aircrafts, energy generating plants, renewable energy systems, etc. (Bonney *et al.*, 2022). Considering the engineering applications, a DT has four main components (Gardner *et al.*, 2020):

- Models (both physics- and data-based),
- Data,
- Digital connectivity, and
- Knowledge.

Generally, DT is linked to physical devices via Internet-of-Things (IoT) applications, predominantly as a technique for retrieving and transferring data (Gilchrist, 2016). Thus, DT is nothing but a the two-way dynamic mapping between a physical object and its virtual model (Angrish *et al.*, 2017). The objective of this virtual ''twinning'' is to allow a well-organized execution of product design and development, manufacturing, maintenance, and numerous other tasks throughout the product lifecycle. Fig.6 provides an overview of the concept of digital twin technology.

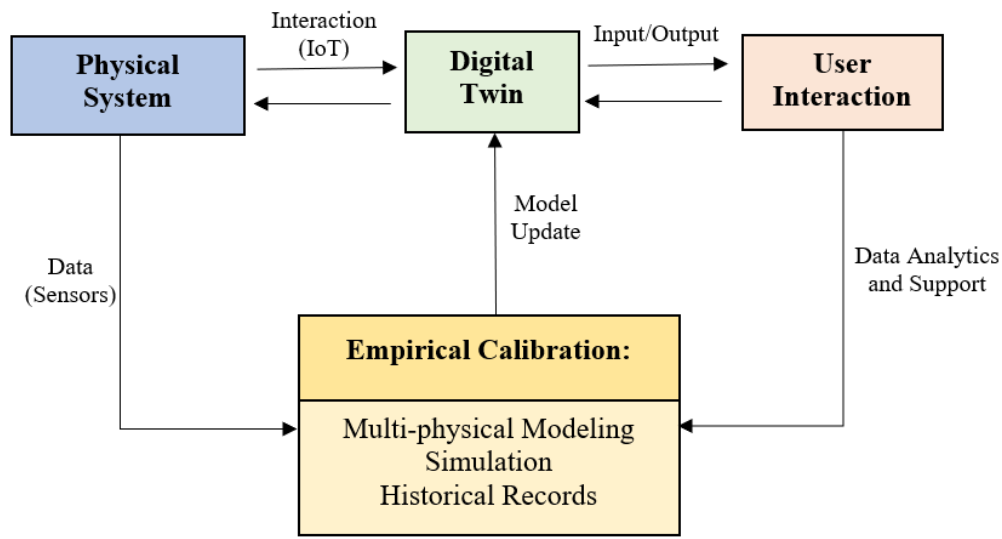

**Fig.6.** Concept of DT technology

The researchers have identified three different application areas for digital twins: manufacturing, healthcare, and smarty city environments (Fuller *et al.*, 2020). For the upcoming industrial 4.0 revolution, industries that don't adopt this technology might be fallen off, nuclear power industry being no exception to that formula. It is considered that digital twin technology is the crucial system for extending nuclear power applications. To preserve world's climate, the role of carbon-free, clean energy sources like nuclear is undeniable (Wang *et al.*, 2011). And to obtain better efficiency and safety, digital twin technology has no substitute. The influence of such digital twin technology might be far-reaching for the nuclear industry (Kochunas and Huan, 2021).

Although many may not believe, DT concept is actually an old concept in the nuclear power industry. In the U.S. Govt. research programs for development of nuclear power reactors, physical twin assets were used (Kochunas and Huan, 2021). Since then, however, the concept of twinning has not evolved that much in this sector. Although there has been a good start on implementation of DT technology, the main limitations that are slowing down the progress are lack of reliable interaction mechanisms between the virtual models and physical objects, the absence of proven reference technologies, and the lack of clear direction of application of DT (Kochunas and Huan, 2021), with the existing models focusing heavily on operation and maintenance (Fuller *et al.*, 2020; Rasheed, San and Kvamsdal, 2020). There are also some unique challenges waiting for a DT in the nuclear field such as the standardization, deciding the best possible way to utilize the existing modeling and simulation options, and the integration of simulation schemes to risk assessment (Kochunas and Huan, 2021).

Since DTs are data-driven and will extensively use computational methods, the concept of using surrogate models in developing DTs is quite compelling. Researchers are exploring the prospect of utilizing surrogate models within DT technology (Chakraborty, Adhikari and Ganguli, 2021). The authors are also on the path of developing a surrogate modeling tool that are dedicated to the DT system for accident tolerant fuels (ATFs). The project is based on the General Atomics Electromagnetic Systems proposal for the DOE Nuclear Power Program Technologies (Jacobsen, 2022). Fig.7 presents the interaction pattern between the physical assets and digital twin (Kochunas and Huan, 2021). However, developing an accurate DT requires robustness against data uncertainty (Pepper, Montomoli and Sharma, 2019; Kumar, Alam, Vučinić, *et al.*, 2020; Kumar, Koutsawa, *et al.*, 2020; Kumar *et al.*, 2022). The authors of this chapter have proposed machine learning (ML)-based uncertainty quantification (UQ) techniques for this purpose (Kumar *et al.*, 2019, 2021; Kumar, Alam, Sjöstrand, *et al.*, 2020). Several new concepts are also being explored (Kobayashi *et al.*, 2022; Kumar *et al.*, 2022).

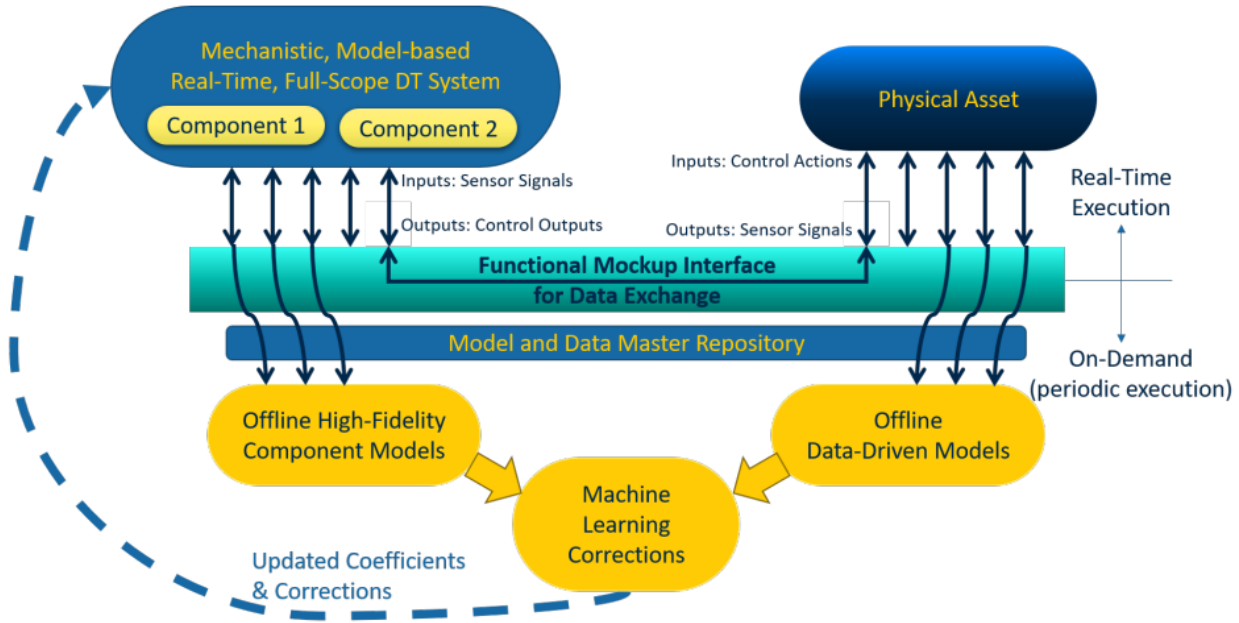

**Fig.7.** Interactions pattern digital twin and physical assets (Kochunas and Huan, 2021)

## 5. Uncertainty quantification

In physical science, the term uncertainty can be defined as the deviation of a measurement, experimental or numerical, from the actual value. Uncertainty quantification (UQ) is the science of characterizing, quantifying, and minimizing uncertainties associated with experiments, numerical models, and algorithms (Kochunas and Huan, 2021). Although the field of quantification of uncertainties is as old as the disciplines of statistics and probability, the systematic assessment of uncertainties and errors and examination of their mode of propagation in models, simulations, and experiments is a newer field of research. Whether the uncertainty is intrinsic to the model or due to the lack of knowledge, uncertainties can generally be classified into aleatoric and epistemic or systemic uncertainty (Helton *et al.*, 2010). Aleatoric uncertainty is the stochastic uncertainty inherent to a quantity of interest or the model and cannot be reduced by more knowledge or training data. Epistemic uncertainty, on the other hand, is the uncertainty related to the oversimplification of the assumptions, missing physics, or lack of enough knowledge, which can be minimized by more appropriate constraints and training data. However, the primary purpose of all uncertainty quantification is the same – to understand the variation of the model output with the original data. Zhang, Yin and Wang (2020) demonstrated the basic framework of uncertainty quantification for any model (as shown in Fig.8).

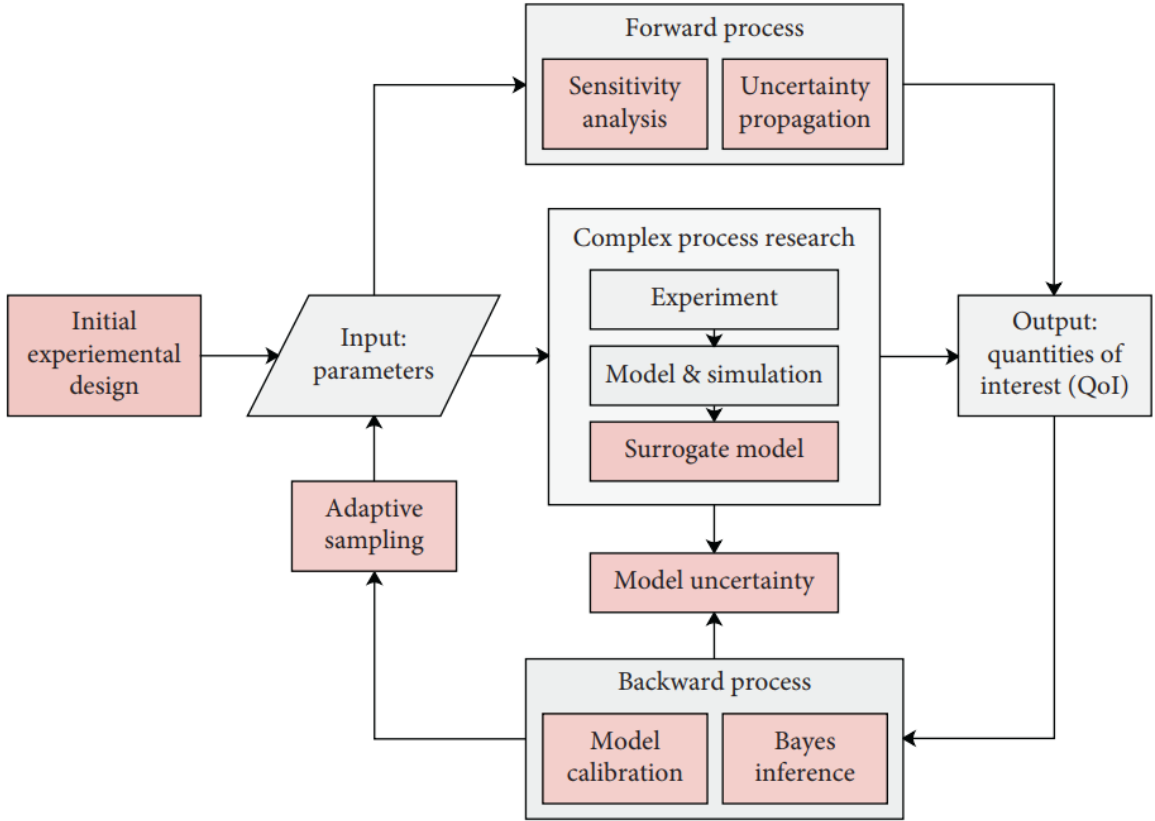

**Fig.8.** UQ framework with main processes (Zhang, Yin and Wang, 2020)

Some popular UQ approaches like polynomial chaos, Gaussian process, Perturbation method, Monte Carlo, neural network, sensitivity analysis, dimension reduction, and inverse uncertainty quantification methods are addressed in the following sections.

### 5.1. Polynomial chaos method

The Hermite polynomial and homogenous chaos methodology were first presented by Wiener (1938). However, it has recently got a new impulse in the last twenty years in the field of uncertainty quantification and has also been modified for both intrusive and non-intrusive use (Ebiwonjumi and Lee, 2021). The basic polynomial chaos equation can be summed up by,

$$Y(x, \delta) = \sum_{i=1}^{n} c_i(x)\, \psi_i(\delta) \qquad (1)$$

Here, Y is the uncertain parameter, $c_i$ denotes the polynomial chaos coefficients, and $\Psi_i$ is a polynomial function of the random variables δ. The variance of Y can be obtained from equation (2).

$$\sigma_Y^2 = E[(Y - E[Y])^2] \qquad (2)$$

Ebiwonjumi and Lee (2021) used the polynomial chaos expansion-based UQ method for efficient computation. Kumar, Alam, Vučinić *et al.* (2020) developed a sparse polynomial chaos expansion-based UQ method for incorporating with digital twins.

## 5.2. Monte Carlo method

The Monte Carlo method is used to approximate a random variable x and, in our case, the uncertainty. Here, the values of $x_1$, $x_2$, …, $x_n$ are generated without any bias from the distribution of X and their average is taken such that,

$$Y(x) = \frac{1}{n} \sum_{i=1}^{n} x_i \qquad (3)$$

With a sample set large enough, this approximation of average value becomes close to the actual average, which would be computationally much expensive to compute otherwise. And then, a close approximation of variance can be computed as,

$$\varphi_Y^2 = E[(Y - E[Y])^2] \qquad (4)$$

Although comparatively less expensive than the frequentist methods, this approach is still very much expensive due to a large sample size requirement for a good approximation. Modern Monte Carlo variants such as the multilevel Monte Carlo method (Koley, Ray and Sarkar, 2021), multi-model Monte Carlo method (Zhang and Shields, 2018), and multi-fidelity Monte Carlo method (Ng and Willcox, 2014) have been developed in order to reduce computational expenses.

## 5.3. Neural Network method

A neural network (NN) or Artificial neural network (ANN) is a black box type approach that mimics the behavior of a human brain where each node acts like a single neuron. A neural network works based on the multilayer perception with one input layer, one or more hidden layers, and one output layer. Each neuron has a weight and a bias value that are to be trained. To incorporate non-linearity into the model, several activation functions like tanh, sigmoid, and ReLU are used (Sharma, Sharma and Athaiya, 2017). Tripathy and Bilionis (2018) mentioned that a real-world problem has a vast number of input parameters in many instances, which makes other uncertainty quantification methods infeasible. The authors, therefore, proposed and demonstrated the efficacy of a deep neural network for such tasks. Hirschfeld *et al.* (2020) performed a model evaluation for several neural network approaches and found that no single model is unequivocally superior in all aspects. Feng, Grana and Balling (2021) used a convolutional neural network for uncertainty analysis in reservoir characterization by segmentation of faults based on seismic images.

## 5.4. Sensitivity analysis method

There are two types of sensitivity analysis (SA) methods: local sensitivity analysis and global sensitivity analysis. Most local SA methods are based on the Taylor series approximation of the considered framework for small perturbations (Helton and Davis, 2003). Global SA, on the other hand, breaks down the overall variance of the model output into contributions of each input parameter, employing several methodologies (Chen, Jin and Sudjianto, 2004; Saltelli *et al.*, 2008). Although most of the global SA models examine the whole extent of all the input variables, it is computationally expensive and indifferent to the interdependency of the input parameters. Recently, in numerous articles, global sensitivity analysis is estimated by calculating Sobol indices from polynomial chaos expansion (Kumar et al, 2020).

## 5.5. Dimension reduction method

In some cases, only some input parameters dictate most part of the variability in a model, while in other cases, all the input variables contribute to the uncertainty. Dimension reduction models quantify uncertainty in these cases in two ways: by reducing the number of probabilistic variables or by transforming the data into lower dimensionality. In reducing the number of probabilistic variables, variables that contribute little to none to the output uncertainty are usually deducted from the model while analyzing uncertainty (Safta *et al.*, 2016). Sometimes the model can be reduced by transforming the model from a higher dimension space to a lower dimension space. Sure, some information is lost in the process, but in many cases, it can still contain most of the model uncertainty (Galbally *et al.*, 2010; Chen and Quarteroni, 2015). Also, reducing a model to a lower order prevents overfitting of the model data (Safta *et al.*, 2016). For almost all the applications, dimension reduction is based on eigenvalue problems. Principal Component Analysis (PCA), Karhunen-Loeve Expension (KLE), Singular Value Decomposition (SVD) are some of the very frequent terms used in model reduction. In these approaches, the eigenvalues of the system are sorted in decreasing order and information related to some of the largest eigenvalues is considered for the analysis.

## 5.6. Inverse uncertainty quantification

For a long time, uncertainty quantification in the nuclear community usually meant forward uncertainty quantification (FUQ), where the flow of information is from input to output. However, another mode of UQ, inverse UQ (IUQ), where information flow from output to input, has been gaining traction recently as it has some advantages over FUQ (Wu *et al.*, 2021). In IUQ, the input uncertainty is measured based on experimental data, whereas FUQ requires past knowledge of the input uncertainty obtained by personal judgment. Wu et al. has presented a survey of IUQ methodologies that can be applied in nuclear thermal hydraulics codes like TRACE and RELAP5 (Wu *et al.*, 2017, 2021).

## 6. Uncertainty Quantification and Nuclear Industry

Being data-driven, surrogate modeling requires the prediction of uncertainty more than a physics-based model as the training of a surrogate model ultimately depends on minimizing cost, the deviation of the model prediction from actual data. The nuclear industry has also seen its fair share of surrogate modeling in the booming of data-driven approaches. Surrogate models have reached from modeling the nuclear reactor core to managing nuclear waste and everything in between. In every surrogate model, uncertainty quantification is critical for understanding the performance of the model. However, there have been numerous instances where the authors just performed surrogate modeling for nuclear systems without considering its uncertainties (Raza and Kim, 2007; Dong *et al.*, 2021). However, in the last few decades, a number of stochastic uncertainty quantification methods such as polynomial chaos (Mathelin, Hussaini and Zang, 2005; Kumar, Alam, Vučinić, *et al.*, 2020), Monte Carlo methods (Fishman, 2013; Koley, Ray and Sarkar, 2021), perturbation method (Taylor III *et al.*, 2003), and sensitivity methods (Zimmerman, Hanson and Davis, 1991) have been proposed.

Yankov (2015) performed an analysis of reactor simulation using the surrogate model. The author, after performing uncertainty quantification, proposed that the reduced-order model containing all one-dimensional elements is sufficiently accurate, although the actual model has a higher dimension. Krivtchik *et al.* (2015) proposed a new approach for uncertainty propagation based on the depletion of the surrogate model in the COSI-S scheme and tested the method on a scenario modeling for the French historical PWR fleet. Presumably, the first direct comparison of the frequentist approach and Bayesian approach was done by King *et al.* (King *et al.*, 2019). The authors showed that for nuclear reaction, the Bayesian approach presents the uncertainties more accurately than the standard frequentist approach. Radaideh and Kozlowski (2019) considered a Gaussian process-based approach for uncertainty quantification in a surrogate model for identifying nuclear fuel performance problems. Figueroa and Göttsche (2021) applied the root-mean-square-metric to measure the uncertainties in a Gaussian process (GP) regression for the surrogate model of discharged fuel nuclide compositions. They have compared the Monte Carlo (MC) method with the GP method and concluded that although the GP regression takes longer than the MC method, it results in a smaller interpolation error. Miles *et al.* (2021) used the Monte Carlo n-particle (MCNP) surrogate model for a similar purpose – radiation source localization. The authors suggested that the performance of the surrogate model can be further optimized by training the surrogate further using the uncertainty quantification provided by the MCNP code.From the above literature survey, it is clearly identified that although there have been a few works here and there, the research exploring UQ methods for surrogate models of nuclear systems is somewhat inadequate. Although there has been a renewed interest among contemporary researchers, a greater emphasis on uncertainty quantification is expected for the successful implementation of surrogate modeling in the nuclear industry.

## 7. Conclusions

Surrogate models introduce a new domain of modern computational research. Supported by deep transfer learning techniques, these models are significantly less expensive computationally compared to conventional physics-based simulation models. Because of being robust, fast, and accurate, they are slowly becoming an integral part of the vision of Industry 4.0. To develop real-time networks for cyber-physical systems and digital twins, surrogate modeling can play a critical role. However, the intrinsic uncertainties within the surrogate models may pose real threat to the accuracy of these models and subsequently obstruct their practical implementation.

This chapter briefly explained how Industry 4.0 and nuclear power sector can be mutually benefited by each other. The importance of computer simulation in the concept of Industry 4.0 was also introduced. This study also reviewed some of the commonly used UQ methods and their applications specific to surrogate models for nuclear systems. From the literature review, it is evident that the number of research works that address the uncertainties associated with surrogate models in nuclear systems is quite inadequate. This has limited the utilization of surrogate modeling technique in many sectors of nuclear engineering. Unless the intrinsic uncertainties of surrogate models for nuclear systems are addressed and properly quantified, the implementation of Industry 4.0 in nuclear power sector is impossible. Therefore, uncertainty quantification of a surrogate model is just as important as the model itself.

## Reference

Abedini, M. and Zhang, C. (2021) 'Performance assessment of concrete and steel material models in ls-dyna for enhanced numerical simulation, a state of the art review', *Archives of Computational Methods in Engineering*, 28(4), pp. 2921–2942.

Angrish, A. *et al.* (2017) 'A flexible data schema and system architecture for the virtualization of manufacturing machines (VMM)', *Journal of Manufacturing Systems*, 45, pp. 236–247. doi: 10.1016/J.JMSY.2017.10.003.

Avvakumov, A. V. *et al.* (2018) 'State change modal method for numerical simulation of dynamic processes in a nuclear reactor', *Progress in Nuclear Energy*, 106, pp. 240–261. doi: 10.1016/J.PNUCENE.2018.02.027.

Bonney, M. S. *et al.* (2022) 'Development of a digital twin operational platform using Python Flask', *Data-Centric Engineering*, 3.

Brereton, R. G. and Lloyd, G. R. (2010) 'Support Vector Machines for classification and regression', *Analyst*, 135(2), pp. 230–267. doi: 10.1039/B918972F.

Chakraborty, S. (2021) 'Transfer learning based multi-fidelity physics informed deep neural network', *Journal of Computational Physics*, 426, p. 109942. doi: 10.1016/J.JCP.2020.109942.

Chakraborty, S., Adhikari, S. and Ganguli, R. (2021) 'The role of surrogate models in the development of digital twins of dynamic systems', *Applied Mathematical Modelling*, 90, pp. 662–681. doi: 10.1016/J.APM.2020.09.037.

Chen, P. and Quarteroni, A. (2015) 'A new algorithm for high-dimensional uncertainty quantification based on dimension-adaptive sparse grid approximation and reduced basis methods', *Journal of Computational Physics*, 298, pp. 176–193. doi: 10.1016/J.JCP.2015.06.006.

Chen, W., Jin, R. and Sudjianto, A. (2004) 'Analytical variance-based global sensitivity analysis in simulation-based design under uncertainty', in *International Design Engineering Technical Conferences and Computers and Information in Engineering Conference*, pp. 953–962.

Clement, A. *et al.* (2021) 'Bayesian Approach for Multigamma Radionuclide Quantification Applied on Weakly Attenuating Nuclear Waste Drums', *IEEE Transactions on Nuclear Science*, 68(9), pp. 2342–2349.

Davis, S. E., Cremaschi, S. and Eden, M. R. (2018) 'Efficient surrogate model development: impact of sample size and underlying model dimensions', in *Computer Aided Chemical Engineering*. Elsevier, pp. 979–984.

Dong, G. *et al.* (2021) 'Deep learning based surrogate models for first-principles global simulations of fusion plasmas', *Nuclear Fusion*, 61(12), p. 126061.

Eason, J. and Cremaschi, S. (2014) 'Adaptive sequential sampling for surrogate model generation with artificial neural networks', *Computers & Chemical Engineering*, 68, pp. 220–232.

Ebiwonjumi, B. and Lee, D. (2021) 'Bayesian method and polynomial chaos expansion based inverse uncertainty quantification of spent fuel using decay heat measurements', *Nuclear Engineering and Design*, 378, p. 111158. doi: 10.1016/J.NUCENGDES.2021.111158.

Emmert-Streib, F. *et al.* (2020) 'An introductory review of deep learning for prediction models with big data', *Frontiers in Artificial Intelligence*, 3, p. 4.

Evins, R. (2013) 'A review of computational optimisation methods applied to sustainable building design', *Renewable and Sustainable Energy Reviews*, 22, pp. 230–245. doi: 10.1016/J.RSER.2013.02.004.

Feng, R., Grana, D. and Balling, N. (2021) 'Uncertainty quantification in fault detection using convolutional neural networks', *Geophysics*, 86(3), pp. M41–M48.

Figueroa, A. and Göttsche, M. (2021) 'Gaussian processes for surrogate modeling of discharged fuel nuclide compositions', *Annals of Nuclear Energy*, 156, p. 108085. doi: 10.1016/J.ANUCENE.2020.108085.

Fishman, G. (2013) *Monte Carlo: concepts, algorithms, and applications*. Springer Science & Business Media.

Fuller, A. *et al.* (2020) 'Digital twin: Enabling technologies, challenges and open research', *IEEE access*, 8, pp. 108952–108971.

Galati, M. and Iuliano, L. (2018) 'A literature review of powder-based electron beam melting focusing on numerical simulations', *Additive Manufacturing*, 19, pp. 1–20. doi: 10.1016/J.ADDMA.2017.11.001.

Galbally, D. *et al.* (2010) 'Non-linear model reduction for uncertainty quantification in large-scale inverse problems', *International journal for numerical methods in engineering*, 81(12), pp. 1581–1608.

Gardner, P. *et al.* (2020) 'Towards the development of an operational digital twin', *Vibration*, 3(3), pp. 235–265.

Gawand, H. L., Bhattacharjee, A. K. and Roy, K. (2017) 'Securing a Cyber Physical System in Nuclear Power Plants Using Least Square Approximation and Computational Geometric Approach', *Nuclear Engineering and Technology*, 49(3), pp. 484–494. doi: 10.1016/J.NET.2016.10.009.

Gilchrist, A. (2016) *Industry 4.0: the industrial internet of things*. Springer.

Goswami, S. *et al.* (2020) 'Transfer learning enhanced physics informed neural network for phase-field modeling of fracture', *Theoretical and Applied Fracture Mechanics*, 106, p. 102447. doi: 10.1016/J.TAFMEC.2019.102447.

Gunal, M. M. (2019) 'Simulation and the fourth industrial revolution', in *Simulation for Industry 4.0*. Springer, pp. 1–17.

Hastie, T., Tibshirani, R. and Friedman, J. (2001) 'Linear methods for regression. the elements of statistical learning: data mining', in *Conference and prediction. Springer Series in Statistics*.

Hedayat, A. *et al.* (2009) 'Estimation of research reactor core parameters using cascade feed forward artificial neural networks', *Progress in Nuclear Energy*, 51(6–7), pp. 709–718. doi: 10.1016/J.PNUCENE.2009.03.004.

Helton, J. C. *et al.* (2010) 'Representation of analysis results involving aleatory and epistemic uncertainty', *International Journal of General Systems*, 39(6), pp. 605–646.

Helton, J. C. and Davis, F. J. (2003) 'Latin hypercube sampling and the propagation of uncertainty in analyses of complex systems', *Reliability Engineering & System Safety*, 81(1), pp. 23–69. doi: 10.1016/S0951-8320(03)00058-9.

Hirschfeld, L. *et al.* (2020) 'Uncertainty quantification using neural networks for molecular property prediction', *Journal of Chemical Information and Modeling*, 60(8), pp. 3770–3780.

Jacobsen, G. (2022) *On the Path to a Nuclear Fuel Digital Twin: Modeling and Simulation of Silicon Carbide Cladding for Accelerated Fuel Qualification*, *US Department of Energy*. Available at: https://www.energy.gov/sites/default/files/2021-11/ne-abstract-silicon-112321.pdf.

Jiang, P., Zhou, Q. and Shao, X. (2020) 'Surrogate Model-Based Engineering Design and Optimization'. doi: 10.1007/978-981-15-0731-1.

Jiang, S. *et al.* (2006) 'Estimation of distribution algorithms for nuclear reactor fuel management optimisation', *Annals of Nuclear Energy*, 33(11–12), pp. 1039–1057. doi: 10.1016/J.ANUCENE.2006.03.012.

King, G. B. *et al.* (2019) 'Direct comparison between Bayesian and frequentist uncertainty quantification for nuclear reactions', *Physical Review Letters*, 122(23), p. 232502.

Kobayashi, K. *et al.* (2022) 'Digital Twin and Artificial Intelligence Framework for Composite Accident-Tolerant Fuel for Advanced Nuclear Systems', in *Hnadbook of Smart Energy Systems*. Springer Nature.

Kochunas, B. and Huan, X. (2021) 'Digital twin concepts with uncertainty for nuclear power applications', *Energies*, 14(14), p. 4235.

Koley, U., Ray, D. and Sarkar, T. (2021) 'Multilevel Monte Carlo Finite Difference Methods for Fractional Conservation Laws with Random Data', *SIAM/ASA Journal on Uncertainty Quantification*, 9(1), pp. 65–105.

Krivtchik, G. *et al.* (2015) 'Analysis of uncertainty propagation in scenario studies: surrogate models application to the French historical PWR fleet', *GLOBAL 2015, Paris, France*.

Kulis, B., Saenko, K. and Darrell, T. (2011) 'What you saw is not what you get: Domain adaptation using asymmetric kernel transforms', in *CVPR 2011*. IEEE, pp. 1785–1792.

Kumar, D. *et al.* (2019) 'Influence of nuclear data parameters on integral experiment assimilation using Cook's distance', in *EPJ Web of Conferences*. EDP Sciences, p. 7001.

Kumar, D., Koutsawa, Y., *et al.* (2020) 'Efficient uncertainty quantification and management in the early stage design of composite applications', *Composite Structures*, 251, p. 112538. doi:

10.1016/J.COMPSTRUCT.2020.112538.

Kumar, D., Alam, S. B., Sjöstrand, H., *et al.* (2020) 'Nuclear data adjustment using Bayesian inference, diagnostics for model fit and influence of model parameters', in *EPJ Web of Conferences*. EDP Sciences, p. 13003.

Kumar, D., Alam, S. B., Vučinić, D., *et al.* (2020) 'Uncertainty quantification and robust optimization in engineering', in *Advances in Visualization and Optimization Techniques for Multidisciplinary Research*. Springer, pp. 63–93.

Kumar, D. *et al.* (2021) 'Quantitative risk assessment of a high power density small modular reactor (SMR) core using uncertainty and sensitivity analyses', *Energy*, 227, p. 120400.

Kumar, D. *et al.* (2022) 'Multi-criteria decision making under uncertainties in composite materials selection and design', *Composite Structures*, 279, p. 114680.

Li, W. *et al.* (2013) 'Learning with augmented features for supervised and semi-supervised heterogeneous domain adaptation', *IEEE Transactions on Pattern analysis and machine intelligence*, 36(6), pp. 1134–1148.

Lu, C. *et al.* (2020) 'Nuclear power plants with artificial intelligence in industry 4.0 era: Top-level design and current applications—A systemic review', *IEEE Access*, 8, pp. 194315–194332.

Mandal, S. *et al.* (2017) 'Nuclear power plant thermocouple sensor-fault detection and classification using deep learning and generalized likelihood ratio test', *IEEE Transactions on nuclear science*, 64(6), pp. 1526–1534.

Mathelin, L., Hussaini, M. Y. and Zang, T. A. (2005) 'Stochastic approaches to uncertainty quantification in CFD simulations', *Numerical Algorithms*, 38(1), pp. 209–236.

McBride, K. and Sundmacher, K. (2019) 'Overview of surrogate modeling in chemical process engineering', *Chemie Ingenieur Technik*, 91(3), pp. 228–239.

Miles, P. R. *et al.* (2021) 'Radiation source localization using surrogate models constructed from 3-D Monte Carlo transport physics simulations', *Nuclear Technology*, 207(1), pp. 37–53.

Nam, J. *et al.* (2017) 'Heterogeneous defect prediction', *IEEE Transactions on Software Engineering*, 44(9), pp. 874–896.

Von Neumann, J. and Goldstine, H. H. (1947) 'Numerical inverting of matrices of high order', *Bulletin of the American Mathematical Society*, 53(11), pp. 1021–1099.

Ng, L. W. T. and Willcox, K. E. (2014) 'Multifidelity approaches for optimization under uncertainty', *International Journal for numerical methods in Engineering*, 100(10), pp. 746–772.

Peng, B. Sen *et al.* (2018) 'Research on intelligent fault diagnosis method for nuclear power plant based on correlation analysis and deep belief network', *Progress in Nuclear Energy*, 108, pp. 419–427. doi: 10.1016/J.PNUCENE.2018.06.003.

Pepper, N., Montomoli, F. and Sharma, S. (2019) 'Multiscale Uncertainty Quantification with Arbitrary Polynomial Chaos', *Computer Methods in Applied Mechanics and Engineering*, 357, p. 112571. doi: 10.1016/J.CMA.2019.112571.

Pereira, A. C. and Romero, F. (2017) 'A review of the meanings and the implications of the Industry 4.0 concept', *Procedia Manufacturing*, 13, pp. 1206–1214. doi: 10.1016/J.PROMFG.2017.09.032.

Radaideh, M. I. and Kozlowski, T. (2019) 'Combining simulations and data with deep learning and uncertainty quantification for advanced energy modeling', *International Journal of Energy Research*, 43(14), pp. 7866–7890. doi: 10.1002/ER.4698.

Rasheed, A., San, O. and Kvamsdal, T. (2020) 'Digital twin: Values, challenges and enablers from a modeling perspective', *Ieee Access*, 8, pp. 21980–22012.

Raza, W. and Kim, K.-Y. (2007) 'Evaluation of surrogate models in optimization of wire-wrapped fuel assembly', *Journal of nuclear science and technology*, 44(6), pp. 819–822.

Razavi, S., Tolson, B. A. and Burn, D. H. (2012) 'Review of surrogate modeling in water resources', *Water Resources Research*, 48(7).

Saeed, H. A. *et al.* (2020) 'Novel fault diagnosis scheme utilizing deep learning networks', *Progress in Nuclear Energy*, 118, p. 103066. doi: 10.1016/J.PNUCENE.2019.103066.

Safta, C. *et al.* (2016) 'Efficient uncertainty quantification in stochastic economic dispatch', *IEEE Transactions on Power Systems*, 32(4), pp. 2535–2546.

Saltelli, A. *et al.* (2008) *Global sensitivity analysis: the primer*. John Wiley & Sons.

Sharma, Sagar, Sharma, Simone and Athaiya, A. (2017) 'Activation functions in neural networks', *towards data science*, 6(12), pp. 310–316.

Sobes, V. *et al.* (2021) 'AI-based design of a nuclear reactor core', *Scientific reports*, 11(1), pp. 1–9.

Sobester, A., Forrester, A. and Keane, A. (2008) *Engineering design via surrogate modelling: a practical guide*. John Wiley & Sons.

Sun, G. and Wang, S. (2019) 'A review of the artificial neural network surrogate modeling in aerodynamic design', *Proceedings of the Institution of Mechanical Engineers, Part G: Journal of Aerospace Engineering*, 233(16), pp. 5863–5872.

Tallman, A. E. *et al.* (2021) 'Surrogate modeling of viscoplasticity in steels: Application to thermal, irradiation creep and transient loading in HT-9 cladding', *JOM*, 73(1), pp. 126–137.

Taylor III, A. C. *et al.* (2003) 'Some advanced concepts in discrete aerodynamic sensitivity analysis', *AIAA journal*, 41(7), pp. 1224–1229.

Tian, K. *et al.* (2020) 'Toward the robust establishment of variable-fidelity surrogate models for hierarchical stiffened shells by two-step adaptive updating approach', *Structural and Multidisciplinary Optimization*, 61(4), pp. 1515–1528.

Tian, K. *et al.* (2021) 'Transfer learning based variable-fidelity surrogate model for shell buckling prediction', *Composite Structures*, 273, p. 114285. doi: 10.1016/J.COMPSTRUCT.2021.114285.

Tripathi, D. *et al.* (2021) 'Model based security verification of Cyber-Physical System based on Petrinet: A case study of Nuclear power plant', *Annals of Nuclear Energy*, 159, p. 108306. doi:

10.1016/J.ANUCENE.2021.108306.

Tripathy, R. K. and Bilionis, I. (2018) 'Deep UQ: Learning deep neural network surrogate models for high dimensional uncertainty quantification', *Journal of Computational Physics*, 375, pp. 565–588. doi: 10.1016/J.JCP.2018.08.036.

Wang, C. and Mahadevan, S. (2011) 'Heterogeneous domain adaptation using manifold alignment', in *Twenty-second international joint conference on artificial intelligence*.

Wang, F. *et al.* (2018) 'Nuclear energy 5.0: new formation and system architecture of nuclear power industry in the new IT era', *Acta Automatica Sinica*, 44(5), pp. 922–934.

Wang, R. *et al.* (2011) 'Path towards achieving of China's 2020 carbon emission reduction target—a discussion of low-carbon energy policies at province level', *Energy Policy*, 39(5), pp. 2740–2747.

Weiss, K., Khoshgoftaar, T. M. and Wang, D. (2016) 'A survey of transfer learning', *Journal of Big data*, 3(1), pp. 1–40.

Wiener, N. (1938) 'The homogeneous chaos', *American Journal of Mathematics*, 60(4), pp. 897–936.

Wu, X. *et al.* (2017) 'Inverse uncertainty quantification of TRACE physical model parameters using sparse gird stochastic collocation surrogate model', *Nuclear Engineering and Design*, 319, pp. 185–200. doi: 10.1016/J.NUCENGDES.2017.05.011.

Wu, X. *et al.* (2021) 'A comprehensive survey of inverse uncertainty quantification of physical model parameters in nuclear system thermal–hydraulics codes', *Nuclear Engineering and Design*, 384, p. 111460. doi: 10.1016/J.NUCENGDES.2021.111460.

Xia, H., Li, B. and Liu, J. (2014) 'Research on intelligent monitor for 3D power distribution of reactor core', *Annals of Nuclear Energy*, 73, pp. 446–454. doi: 10.1016/J.ANUCENE.2014.07.033.

Yankov, A. (2015) 'Analysis of Reactor Simulations Using Surrogate Models.'

Yao, Y. *et al.* (2022) 'Model-based Deep Transfer Learning Method to Fault Detection and Diagnosis in Nuclear Power Plants', *Frontiers in Energy Research*, 10, pp. 1–12.

Yao, Y., Wang, J. and Xie, M. (2022) 'Adaptive residual CNN-based fault detection and diagnosis system of small modular reactors', *Applied Soft Computing*, 114, p. 108064. doi: 10.1016/J.ASOC.2021.108064.

Yondo, R. *et al.* (2019) 'A review of surrogate modeling techniques for aerodynamic analysis and optimization: current limitations and future challenges in industry', *Advances in evolutionary and deterministic methods for design, optimization and control in engineering and sciences*, pp. 19–33.

Zhang, J. and Shields, M. D. (2018) 'On the quantification and efficient propagation of imprecise probabilities resulting from small datasets', *Mechanical Systems and Signal Processing*, 98, pp. 465–483. doi: 10.1016/J.YMSSP.2017.04.042.

Zhang, J., Yin, J. and Wang, R. (2020) 'Basic framework and main methods of uncertainty quantification', *Mathematical Problems in Engineering*, 2020.

Zhou, J. *et al.* (2014) 'Hybrid heterogeneous transfer learning through deep learning', in *Proceedings of the AAAI Conference on Artificial Intelligence*.

Zimmerman, D. A., Hanson, R. T. and Davis, P. A. (1991) *A comparison of parameter estimation and sensitivity analysis techniques and their impact on the uncertainty in ground water flow model predictions*. Nuclear Regulatory Commission, Washington, DC (United States).